\documentclass[letterpaper]{article} 
\usepackage{aaai25}  
\usepackage{times}  
\usepackage{helvet}  
\usepackage{courier}  
\usepackage[hyphens]{url}  
\usepackage{graphicx} 
\usepackage{natbib}  
\usepackage{caption} 
\usepackage{upgreek}
\usepackage{amsfonts}
\usepackage{amsthm}
\usepackage{amsmath}
\usepackage{siunitx}

\usepackage[noend]{algpseudocode}
\usepackage{algorithm} 
\usepackage{amsmath}

\usepackage{chngcntr}
\counterwithin{figure}{section}
\counterwithin{table}{section}

\usepackage{newfloat}
\usepackage{listings}
\DeclareCaptionStyle{ruled}{labelfont=normalfont,labelsep=colon,strut=off} 
\floatstyle{ruled}
\newfloat{listing}{tb}{lst}{}
\floatname{listing}{Listing}

\title{Adaptive Sampling to Reduce Epistemic Uncertainty Using Prediction Interval-Generation Neural Networks}
\author{
    Giorgio Morales, John W. Sheppard
}
\affiliations{
    Gianforte School of Computing\\ Montana State University, 
    Bozeman, MT 59717, USA\\
    giorgiomorales@ieee.org; john.sheppard@montana.edu
}

\begin{document}

\maketitle

\begin{abstract}
Obtaining high certainty in predictive models is crucial for making informed and trustworthy decisions in many scientific and engineering domains. 
However, extensive experimentation required for model accuracy can be both costly and time-consuming. 
This paper presents an adaptive sampling approach designed to reduce epistemic uncertainty in predictive models. 
Our primary contribution is the development of a metric that estimates potential epistemic uncertainty leveraging prediction interval-generation neural
networks.
This estimation relies on the distance between the predicted upper and lower bounds and the observed data at the tested positions and their neighboring points. 
Our second contribution is the proposal of a batch sampling strategy based on Gaussian processes (GPs). 
A GP is used as a surrogate model of the networks trained at each iteration of the adaptive sampling process. 
Using this GP, we design an acquisition function that selects a combination of sampling locations to maximize the reduction of epistemic uncertainty across the domain.
We test our approach on three unidimensional synthetic problems and a multi-dimensional dataset based on an agricultural field for selecting experimental fertilizer rates.
The results demonstrate that our method consistently converges faster to minimum epistemic uncertainty levels compared to Normalizing Flows Ensembles, MC-Dropout, and simple GPs.
\end{abstract}


\begin{links}
\small
    \link{Code}{https://github.com/NISL-MSU/AdaptiveSampling}
\end{links}

\section{Introduction}

In various scientific and engineering fields, the development of accurate predictive models frequently relies on experimentation. 
Conducting these experiments can be costly and time-consuming, making it important to adopt strategies that extract the most valuable information from each experiment. 
One notable example is precision agriculture (PA) where experimental results may require an entire growing season to manifest, and only a portion of the field is allocated for such trials~\cite{lawrence_probabilistic_2015}.
This is exacerbated by the fact data can often only be collected every other year, due to crop rotation.

Adaptive sampling (AS) techniques offer a promising solution by selecting samples intelligently that contribute most to improving model accuracy and reducing uncertainty~\cite{di_fiore_active_2023}.
This work focuses on sampling techniques designed to reduce uncertainty in the prediction models across the entire input domain.
Such techniques are essential for enhancing trust in decision-making systems whose optimization processes rely on accurate prediction models.
For instance, in PA, determining optimal fertilizer rates depends on the shape of estimated nitrogen-yield response (N-response) curves~(\citet{bullock94}, \citet{rcurves2023}).
These curves represent the estimated crop yield values at specific field sites in response to all admissible fertilizer rates.
Uncertainty across the domain can severely affect the survey shapes, leading to unreliable recommended fertilizer rates.

We note a distinction between two types of uncertainty: epistemic and aleatoric.
Epistemic uncertainty represents the portion of total uncertainty that can be reduced by gathering more information or improving the prediction model. 
On the other hand, aleatoric uncertainty is the inherent and irreducible component of uncertainty due to the random nature of the data itself~\cite{hullermeier_aleatoric_2021,nguyen_how_2022}.
The total uncertainty associated with a prediction ($\sigma^2_y$) encapsulates both the aleatoric ($\sigma^2_{a}$) and epistemic ($\sigma^2_{e}$) components; i.e., $\sigma^2_y = \sigma^2_a + \sigma^2_e$.
Prediction intervals (PIs) offer a comprehensive representation of this total uncertainty by estimating the upper and lower bounds within which a prediction is expected to fall with a given probability~\cite{LUBE}.

Several methods have been proposed to reduce uncertainty through iterative sampling.
However, the majority of these methods have been developed within the framework of active learning (AL)~\cite{nguyen_epistemic_2019,NFs} or in contexts where the primary objective is to identify the location of local or global optima~\cite{entropy,nguyen_efficient_2019}.   

It is important to note that AS and AL fields do not completely overlap~\cite{di_fiore_active_2023}.
In AL, the objective is to select training data within a limited budget to maximize model performance.
AL can be categorized into population-based AL, where the test input distribution is known, and pool-based AL, where a pool of unlabeled samples is provided.
Our problem configuration does not align with those categories as it is not limited to predefined data pools or known distributions.
Instead, it aims to sample from an open domain continuously, focusing on reducing epistemic uncertainty across the entire input space.

In this paper, we propose a method to reduce epistemic uncertainty through adaptive sampling using PIs generated by neural networks (NNs). 
Our method, called Adaptive Sampling with Prediction-Interval Neural Networks (ASPINN), involves training a dual NN architecture comprising a target-estimation network and a PI-generation network.
The objective of such NNs is to produce high-quality PIs that reflect both aleatoric and epistemic uncertainties. 
Our specific contributions are:
\begin{enumerate}
    \item We introduce a novel metric based on NN-generated PIs to quantify potential levels of epistemic uncertainty.
    \item We present an AS method called ASPINN. At each iteration, it builds a Gaussian Process (GP) from calculated potential epistemic uncertainty levels. The GP, a surrogate for the NN models, estimates potential epistemic uncertainty changes across the domain after sampling specific locations. An acquisition function then uses the GP to select sampling locations, aiming to minimize global epistemic uncertainty throughout the input domain.
    \item We tackle a real-world application and present an AS benchmark problem that focuses on reducing the epistemic uncertainty of an agricultural field site. 
    \item Our method is shown to converge faster to minimum epistemic uncertainty levels than the compared methods.
\end{enumerate}

\section{Related Work}

The problem addressed in this work shares similarities with Bayesian Optimization (BO), where at each iteration, data points are sampled at locations expected to yield significant improvements in the objective function according to a specified acquisition function.
BO methods build a probabilistic model of the objective function, often a GP, to select the most promising points for evaluation~\cite{garnett_bayesian_2023}.

Traditional BO methods explore the domain space sequentially; however, \citet{batchBO} proposed a batch sampling strategy for BO that accounts for the interactions between different evaluations in the batch using a penalized acquisition function.
Some BO strategies focus on maximizing information gain. 
For instance, \citet{MES} introduced an acquisition function called max-value entropy search (MES), which balances exploration of areas with higher uncertainty in the surrogate model and exploitation towards the believed optimum. 
In addition, \citet{nguyen_efficient_2019} presented the predictive variance reduction search (PVRS) strategy, which reduces uncertainty at perceived optimal locations, leading to convergence when uncertainty at all perceived optimal locations is minimized.

In typical BO applications, the objective is to identify a single location that corresponds to the local or global optimum of an objective function ($\arg\max f(\mathbf{x})$). 
In contrast, the solution to our problem consists of an augmented dataset that yields minimum epistemic uncertainty across the entire input space. 
In the fertilizer rate optimization problem discussed in the previous section, finding the rate that produces the higher estimated yield value does not necessarily coincide with the economic optimum nitrogen rate (EONR).
The EONR is the N rate beyond which there is no actual profit for the farmers and its calculation depends on the shape of the N-response curves~\cite{bullock94}.
Therefore, the epistemic uncertainty across all admissible N rates should be reduced to provide reliable EONR recommendations for future growing seasons.

Similarly, active learning is closely related to this work.
The primary distinction is that AL, given known input distributions (population-based AL) or a set of unlabeled points (pool-based AL), aims to select the minimum number of training examples to maximize model performance~\cite{di_fiore_active_2023}. 
In contrast, our approach is agnostic of the input distribution and is not restricted to a fixed pool of training candidates.
Furthermore, our focus being on reducing uncertainty only considers model prediction improvement as a side-effect.
What is more, it allows for repetitive sampling at a single location.

Despite the distinction above, some AL techniques can be adapted to our problem.
In particular, we are interested in methods that decompose uncertainty into its aleatoric and epistemic components.
A common approach is to use Monte-Carlo Dropout (MC-Dropout)~\cite{pmlr-v48-gal16} to quantify epistemic uncertainty in NNs.
MC-Dropout uses dropout repeatedly to select random subsamples of active nodes in the network, turning a single network into an ensemble.
Hence, epistemic uncertainty is represented by the sample variance of the ensemble predictions.

Furthermore, \citet{VA} used a variance attenuation (VA) loss function to disentangle the epistemic and aleatoric components from the outputs of ensemble models.
However, \citet{onestepcloser} pointed out that VA-based methods overestimate aleatoric uncertainty.
In response, they presented a denoising approach that involves incorporating a variance approximation module into a trained prediction model to identify the aleatoric uncertainty.
Finally, \citet{NFs} proposed using an ensemble of normalizing flows (NFs), created using dropout masks, to estimate both aleatoric and epistemic uncertainty.
To demonstrate their results, they suggested an AL framework that compares various uncertainty estimation methods.
These methods are used to sample multiple-point candidates and select those with the highest epistemic uncertainty.


\section{Proposed Method}

In this work, we examine a system defined by an input vector $\mathbf{x} \in \mathbb{R}^d$ and a scalar response $y \in \mathbb{R}$.  
The system's underlying function $f:\mathcal{X} \rightarrow \mathcal{Y}$ maps the input value space and the response value space such that $y = f(\mathbf{x}) + \varepsilon_a(\mathbf{x})$, where $\varepsilon_a(\mathbf{x})$ is a random variable representing the error term that is a function of the system's aleatoric uncertainty, $\sigma^2_a(\mathbf{x})$.

Let $\mathcal{D}_t = (\mathbf{X}_{obs}^{(t)}, \mathbf{Y}_{obs}^{(t)})$ represent the dataset available at iteration $t$ consisting of $n_t$ observations, where $\mathbf{X}_{obs}^{(t)} = \{\mathbf{x}_1, \dots, \mathbf{x}_{n_t}\}$ and $\mathbf{Y}_{obs}^{(t)} = \{y_1, \dots, y_{n_t}\}$.
A prediction model $\hat{f}_t:\mathcal{X} \rightarrow \mathcal{Y}$ with parameters $\boldsymbol{\theta}_f$ is trained by minimizing the mean squared error of the estimation:
\begin{equation*}
   \min_{\boldsymbol{\theta}_f} \frac{1}{n_t} \sum_{(\mathbf{x}_i, y_i) \in \mathcal{D}_t} (\hat{f}_{t}(\mathbf{x}_i) - y_i)^2. 
\end{equation*}

We aim to identify a batch $\mathbf{X}_{acq}^{(t)} = \{ \mathbf{x}_{t, 1}, \dots, \mathbf{x}_{t, B} \}$ of $B$ recommended sampling locations for the next iteration. 
These locations are chosen to minimize the epistemic uncertainty across the entire input space given a model $\hat{f}_t$ trained on $\mathcal{D}_t$.
The epistemic uncertainty, $\sigma^2_e(\mathbf{x}_p)$, arises from the lack of knowledge about $f$ and is due to the limitations of the prediction model trained on the observed dataset.

\begin{figure}[!t]
    \centering
    \includegraphics[width=0.85\columnwidth]{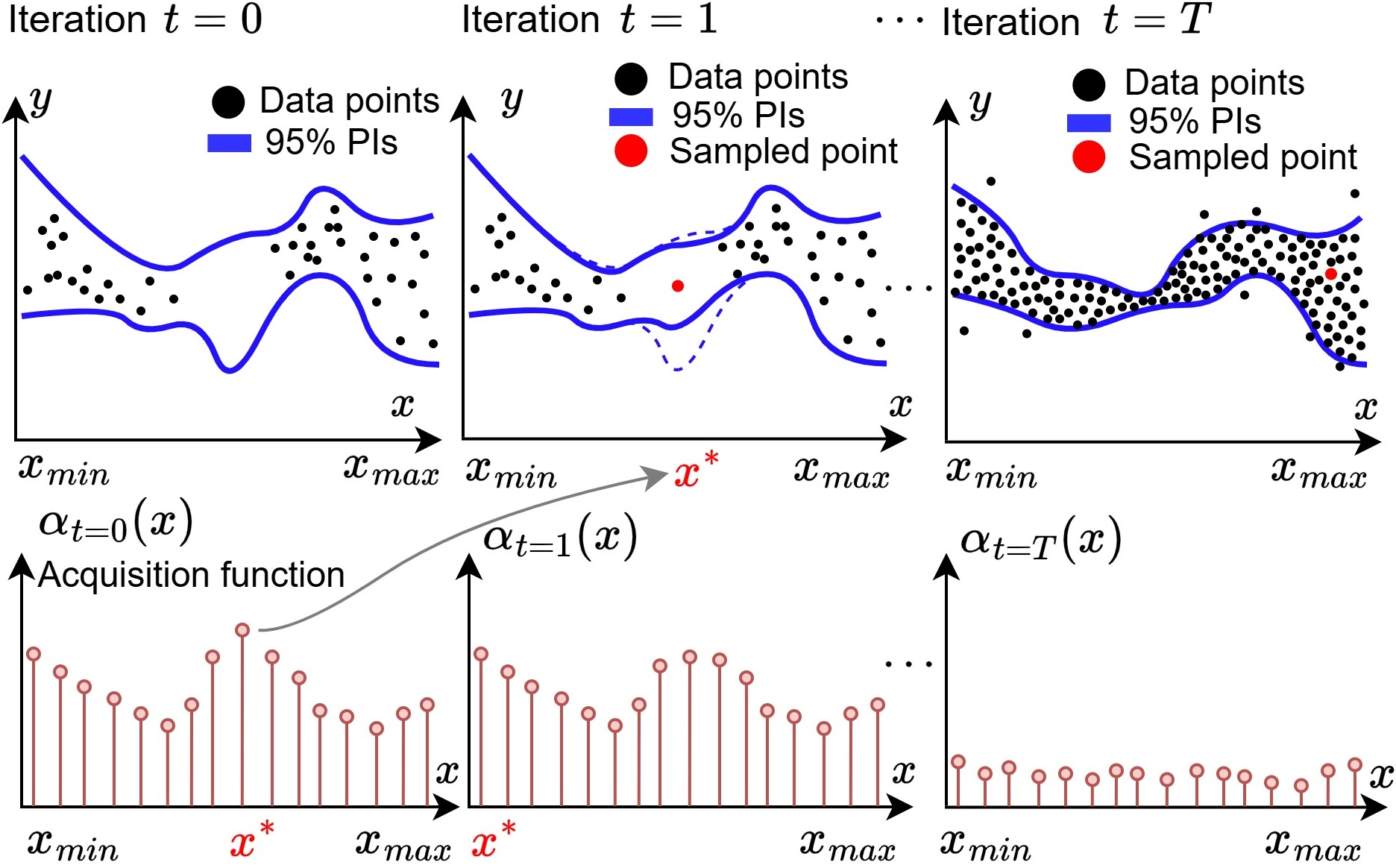}
    \caption{Epistemic uncertainty minimization through AS.}
    \label{fig:framework}
\end{figure}

Preferences over potential sampling locations are encoded by an acquisition function $\alpha_t(\mathbf{x})$.
Suppose $J(\mathcal{D}_t)$ is a function that reflects the total potential epistemic uncertainty across the domain.
Then $\alpha_t(\mathbf{x})$ is designed to reflect the expected decrease in epistemic uncertainty $\mathbb{E}[J(\mathcal{D}_t) - J(\mathcal{D}_{t} \cup  (\mathbf{x}, y) )]$ after making an observation at location $\mathbf{x}$.
Fig.~\ref{fig:framework} depicts an instance of our problem.
Here, $x^*$ represents the selected sampling position at each iteration (i.e., $B=1$).
For the general case where $B>1$, the decision on where to sample the $k$-th element of the batch, $\mathbf{x}_{t, k}$, depends on the estimated effect of the previous $k-1$ samples of the same batch. 
This requires a batch sampling strategy, which will be explored in this paper.

In the following, we describe the components of our ASPINN method.
We lay out the steps to derive a metric that reflects the epistemic uncertainty associated with an input value based on PIs. 
The metric is then used to design an acquisition function that allows for the selection of a batch of sampling locations, which are expected to minimize the global epistemic uncertainty during the next AS iteration.

\subsection{Prediction Interval Generation}

We generate PIs for quantifying the total uncertainty associated with a given sample, thus accounting for both aleatoric and epistemic uncertainty.
We employ an NN-based PI generation method called DualAQD~\cite{DualAQD}.
This method uses two companion NNs: a target-estimation NN and a PI-generation NN, whose computed functions are denoted as $\hat{f}_t(\cdot)$ and $\hat{g}_t(\cdot)$, respectively.
Network $\hat{f}_t(\cdot)$ is trained on $\mathcal{D}_t$ to minimize the target estimation error so that $\hat{y} = \hat{f}_t(\mathbf{x})$ and $\hat{y} \approx y$.
Network $\hat{g}_t(\cdot)$ produces two outputs $[\hat{y}^\ell, \, \hat{y}^u] = \hat{g}_t(\mathbf{x})$, which correspond to the PI lower and upper bounds.
Note that $\hat{g}_t(\mathbf{x})$ makes no assumptions about the underlying uncertainty distribution.

Network $\hat{g}_t(\cdot)$ is trained using the DualAQD loss function to produce high-quality PIs that are as narrow as possible while capturing some specified proportion of the predicted data points (e.g., 95\%).
However, the model should produce wider PIs for out-of-distribution (OOD) samples since these samples are not well-represented in the training set, leading to higher associated epistemic uncertainty.
To address this, the bias weights of $\hat{g}_t(\cdot)$ are initialized to generate wide PIs, similar to the approach proposed by \citet{pi3nn}. 
The rationale is that these bias weights will decrease during training for in-distribution samples, resulting in narrower PIs, but will remain high for OOD samples, ensuring appropriately wider PIs to reflect the increased uncertainty.

\subsection{Potential Epistemic Uncertainty}

Let $\sigma^2_e(\mathbf{x}_p)$ represent the epistemic uncertainty at a certain location $\mathbf{x}_p \in \mathcal{X}$.
The PI lower and upper bounds generated by NN $\hat{g}_t(\cdot)$ at $\mathbf{x}_p$ are denoted as $\hat{y}_t^\ell(\mathbf{x}_p)$ and $\hat{y}_t^u(\mathbf{x}_p)$, respectively.
We claim that using PIs alone does not provide sufficient information to determine $\sigma^2_e(\mathbf{x}_p)$.
Consider $\mathbf{x}_p$ as an OOD sample.
We may state that the total uncertainty associated with $\mathbf{x}_p$ is primarily due to epistemic uncertainty given the lack of knowledge of the prediction model about the system's behavior in this region of the input domain.

However, we cannot estimate the aleatoric uncertainty around $\mathbf{x}_p$ until we gather observations in such domain region.
Alternative methods can be used but they require making assumptions about the noise distribution~\cite{seitzer2022on}, training an ensemble of models~\cite{NFs}, or using additional trainable modules~\cite{onestepcloser}.
Therefore, the total uncertainty conveyed by the interval $[\hat{y}_t^\ell(\mathbf{x}_p), \hat{y}_t^u(\mathbf{x}_p)]$ cannot be split effectively into its epistemic and aleatoric components without further information.

Instead of attempting to provide a metric that accurately estimates $\sigma^2_e(\mathbf{x}_p)$ directly, we propose a metric that reflects the potential levels of epistemic uncertainty.
Let $\mathcal{N}(\mathbf{x}_p) = \{ \mathbf{x} \in \mathbf{X}_{obs}^{(t)} | \, \| \mathbf{x} - \mathbf{x}_p \|_2 \leq \theta\}$ denote a neighborhood that considers all samples whose Euclidian distance to $\mathbf{x}_p$ is less than a hyperparameter threshold $\theta$. 
We create the set of input--response pairs $\mathcal{R}(\mathcal{N}(\mathbf{x}_p)) = \{ (\mathbf{x}, y) \, | \, (\mathbf{x}, y) \in \mathcal{D}_t, \mathbf{x} \in \mathcal{N}(\mathbf{x}_p),  \hat{y}^\ell(\mathbf{x}) \leq y \leq \hat{y}^u(\mathbf{x}) \}$ using the samples in $\mathcal{N}(\mathbf{x}_p)$ whose response values fall within their corresponding PI.
Thus, we present the metric $Q_t(\mathbf{x}_p)$, defined as:
\begin{equation}
\label{eq:Q1}
Q_t(\mathbf{x}_p) \!\! = \!\! 
\begin{cases}
\begin{aligned}
&\min\limits_{\substack{(\mathbf{x}, y) \in  \mathcal{R}(\mathcal{N}(\mathbf{x}_p)) }} (\hat{y}^u(\mathbf{x}) \!-\! y) +\!\\
&\min\limits_{\substack{(\mathbf{x}, y) \in  \mathcal{R}(\mathcal{N}(\mathbf{x}_p)) }} (y \!-\! \hat{y}^\ell(\mathbf{x}))
\end{aligned} & \!\!\!\text{if } \mathcal{N}(\mathbf{x}_p) \!\neq\! \emptyset \\
\hat{y}_t^u(\mathbf{x}_p) - \hat{y}_t^\ell(\mathbf{x}_p) & \!\!\!\text{if } \mathcal{N}(\mathbf{x}_p) \!=\! \emptyset
\end{cases}
\end{equation}

\begin{figure}[!t]
    \centering
    \includegraphics[width=0.75\columnwidth]{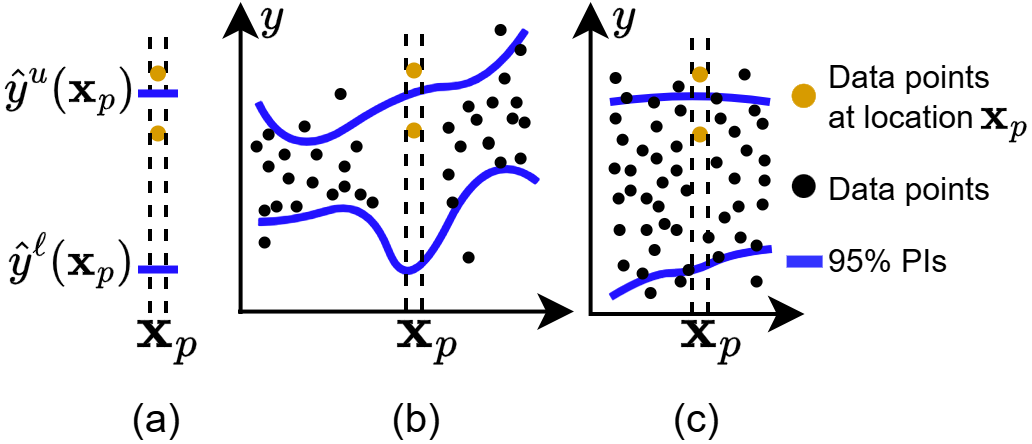}
    \caption{PIs generated at location $\mathbf{x}_p$. \textbf{(a)} Data points located at $\mathbf{x}_p$ only. \textbf{(b)} PI width is affected by epistemic uncertainty. \textbf{(b)} PI width is mainly due to aleatoric uncertainty.}
    \label{fig:PIs}
\end{figure}

The local neighborhood of $\mathbf{x}_p$ may contain important contextual information that an analysis at a single location $\mathbf{x}_p$ cannot capture.
For instance, Fig.\ref{fig:PIs}a illustrates an interval $\mathrm{PI}(\mathbf{x}_p) = [\hat{y}_t^\ell(\mathbf{x}_p), \hat{y}_t^u(\mathbf{x}_p)]$ generated at a single location.
Suppose $Q_t(\mathbf{x}_p)$ is calculated using $\mathrm{PI}(\mathbf{x}_p)$ only (i.e., $\theta = 0$).
Since a single point lies within the interval, $Q_t(\mathbf{x}_p)$ is equal to the PI width, indicating that the epistemic uncertainty at $\mathbf{x}_p$ can potentially be completely reduced.
Fig.\ref{fig:PIs}b depicts a case in which the PI shown in Fig.\ref{fig:PIs}a is located in a region of the domain with low data density.
As such, there exists an epistemic component that entails that the PI width could be reduced by acquiring more data in this region.

Conversely, Fig.\ref{fig:PIs}c shows a similar PI in a high data density context.
Here, a reduction in $\mathrm{PI}(\mathbf{x}_p)$ will also lead to a decrease in the PI widths of adjacent locations, provided that the uncertainty at $\mathbf{x}_p$ is not independent of its surroundings.
However, model $\hat{g}_t(\cdot)$ is trained to produce narrow PIs while maintaining a nominal coverage (e.g., 95\%).
Thus, it will not reduce $\mathrm{PI}(\mathbf{x}_p)$ if this reduction would result in several samples near the PI bounds being excluded from their intervals.
Notice that if $\theta > 0$, then $Q_t(\mathbf{x}_p) \approx 0$, indicating minimal potential epistemic uncertainty around $\mathbf{x}_p$.


\subsection{Batch Sampling}

When multiple locations are sampled at each iteration, decisions for the entire batch are made based on the current model without observing any data from the batch until the next iteration.
Hence, it is necessary to simulate the decisions that would be made under the equivalent sequential policy (i.e., when $B=1$)~\cite{batchBO}.
In other words, the decision of selecting the $k$-th element of the $t$-th batch, $\mathbf{x}_{t, k}$, should incorporate the estimates of change in uncertainty after sampling at locations $\mathbf{x}_{t, 1} , \dots, \mathbf{x}_{t, k-1}$ (i.e., $\mathbf{x}_{t, 1:k-1}$).
Following a greedy sampling strategy, we have:

\begin{equation}
    \label{eq:sample}
    \mathbf{x}_{t, k} = \underset{\mathbf{x}_p \in \mathcal{X}}{\text{argmax}}\  \alpha_t(\mathbf{x}_p \, | \,  \mathbf{x}_{t, 1:k-1}).
\end{equation}

We consider an acquisition function that estimates the reduction in the total potential epistemic uncertainty across the domain when making an observation at a given location $\mathbf{x}_p$:
\[
   \alpha_t(\mathbf{x}_p \, | \,  \mathbf{x}_{t, 1:k-1}) = J \left(\mathcal{D}_{t,k-1} \right) -
   J(\mathcal{D}_{t,k-1} \cup (\mathbf{x}_p, \hat{f}_t(\mathbf{x}_p))  ). 
\]
$\mathcal{D}_{t,k-1}$ is the dataset $\mathcal{D}_{t}$ augmented with the first $k-1$ samples of the batch and their corresponding estimated response values.
The potential epistemic uncertainty at $\mathbf{x}$ during the $t$-iteration after sampling the first $k$ elements of the batch is denoted as $Q_{t,k}(\mathbf{x})$. 
Thus, the total potential epistemic uncertainty is calculated as $J(\mathcal{D}_{t,k}) = \sum_{\mathbf{x} \in \mathcal{X}} Q_{t,k}(\mathbf{x})$, where $J(\mathcal{D}_{t,0}) = J(\mathcal{D}_{t})$ and $Q_{t,0}(\mathbf{x}) = Q_{t}(\mathbf{x})$.

Thus, $J(\mathcal{D}_{t})$ is computed based on $Q_{t}(\mathbf{x})$, which is derived from the outputs produced by NNs $\hat{f}_t(\cdot)$ and $\hat{g}_t(\cdot)$ (Eq.~\ref{eq:Q1}), trained on $\mathcal{D}_t$.
To calculate $J(\mathcal{D}_{t,k-1} \cup (\mathbf{x}_p, \hat{f}_t(\mathbf{x}_p)))$ in a similar manner, it is necessary to train both NNs on the augmented dataset $\mathcal{D}_{t,k-1} \cup (\mathbf{x}_p, \hat{f}_t(\mathbf{x}_p))$. 
According to Eq.~\ref{eq:sample}, this operation would need to be repeated $\forall \mathbf{x}_p \in \mathcal{X}$ and $\forall k \in [1, \dots, B]$ and, as such, becomes impractical.
Therefore, motivated by most BO-based approaches, we use a GP as a surrogate model. 
The objective is to simulate, with low computational cost, how the potential epistemic uncertainty would be affected throughout the entire domain after observing a sample at a given position.

Let us define a GP $p(\hat{f}_t) = \mathcal{GP}(\mu_t, \mathbf{K}_t)$ that serves as a surrogate model for $\hat{f}_t(\cdot)$ and its associated epistemic uncertainty during the $t$-th iteration.
This GP is characterized by the mean function $\mu_t$ and the positive-definite covariance matrix $\mathbf{K}_t$.
Functions $\mu_t$ and $\mathbf{K}_t$ are initialized based on the estimations generated by $\hat{f}_t(\cdot)$ and $\hat{g}_t(\cdot)$, trained on $\mathcal{D}_t$.

For the mean function, we consider $\mu_t(\mathbf{x}) = \hat{f}_t(\mathbf{x})$.
On the other hand, the diagonal elements of $\mathbf{K}_t$ reflect the uncertainty in the predictions $\hat{f}_t(\mathbf{x})$ due to epistemic uncertainty.
Since this uncertainty varies across the domain, it represents heteroscedastic noise. 
Considering that the uncertainty at a given position may be correlated with nearby positions, $\mathbf{K}_t$ is structured as a matrix with non-zero off-diagonal elements.
Thus, the scale of $\mathbf{K}_t$ depends on location and is calculated according to the potential epistemic uncertainty:
\begin{equation*}
  \mathbf{K}_t(\mathbf{x}, \mathbf{x}') =
    \begin{cases}
      Q_t(\mathbf{x}), & \text{if $\mathbf{x} = \mathbf{x}'$}\\
      \rho(\mathbf{x}, \mathbf{x}') \sqrt{Q_t(\mathbf{x}) Q_t(\mathbf{x}')}, & \text{otherwise},
    \end{cases}
 \label{eq:k}
\end{equation*}
where $\rho(\mathbf{x}, \mathbf{x}')$ indicates the correlation between positions $\mathbf{x}$ and $\mathbf{x}'$.
We use the radial basis function (RBF) such that $\rho(\mathbf{x}, \mathbf{x}') = e^{-\frac{\|\mathbf{x} - \mathbf{x}'\|^2}{2r^2} }$, where $r$ is a tunable hyperparameter.

Given we want to assess the impact of observing a data point at a given position $\mathbf{x}_p$,
we condition the GP on the data point $(\mathbf{x}_p, \hat{f}_t(\mathbf{x}_p))$, resulting in a GP posterior $p(\hat{f}_t | (\mathbf{x}_p, \hat{f}_t(\mathbf{x}_p)))$ whose covariate matrix is denoted as $\mathbf{K}_t(\mathbf{x}, \mathbf{x}' \,|\, \mathbf{x}_p)$.
In general, the covariance matrix when sampling the $k$-th element of the batch is denoted as $\mathbf{K}_t(\mathbf{x}, \mathbf{x}' \,|\, \mathbf{x}_{t, 1}, \dots, \mathbf{x}_{t, k})$ and $Q_{t,k} = \mathrm{diag} \left( \mathbf{K}_t(\mathbf{x}, \mathbf{x}' \,|\, \mathbf{x}_{t, 1}, \dots, \mathbf{x}_{t, k}) \right)$. 

Given $\mathbf{x}_p$, the covariance matrix is updated as follows:
\begin{equation*}
\label{eq:covariance}
\begin{split}
    \mathbf{K}_t(\mathbf{x}, \mathbf{x}' \,|\, \mathbf{x}_p) = &\mathbf{K}_t(\mathbf{x}, \mathbf{x}') - \\
    &\mathbf{K}_t(\mathbf{x}, \mathbf{x}_p) \mathbf{K}_t(\mathbf{x}_p, \mathbf{x}_p) ^{-1} \mathbf{K}_t(\mathbf{x}_p, \mathbf{x}').
\end{split}
\end{equation*}
Hence, the updated GP variance at $\mathbf{x}_p$ collapses to zero after observing a data point at that position.
Note that this would only happen when $Q_t(\mathbf{x}_p)$ reflects the level of epistemic uncertainty exclusively.
In practice, this assumption may not hold.
Nevertheless, it allows us to construct a heuristic that guides the search toward locations where new observations would potentially cause the greatest uncertainty reduction.  
The next sampling location is selected using Eq.~\ref{eq:sample} based on the total potential epistemic uncertainty after observing a data point at $\mathbf{x}_p$, which is given by:
\begin{equation*}
    \label{eq:select}
    J\left(\mathcal{D}_{t} \cup (\mathbf{x}_p, \hat{f}_t(\mathbf{x}_p)) \right) = \sum \mathrm{diag} \left(\mathbf{K}_t(\mathbf{x}, \mathbf{x}' \,|\, \mathbf{x}_p)\right).
\end{equation*}

\section{Experimental Results}

We compared ASPINN to three methods adapted for AS: Normalizing flows ensembles (NF-Ensemble)~\cite{NFs}, a standard GP~\cite{Gpytorch}, and MC-Dropout~\cite{pmlr-v48-gal16}.
For our experiments, we considered three synthetic one-dimensional (1-D) regression problems and one multidimensional regression problem based on a real-world problem.
We used synthetic problems given that, in AS, we are required to sample at locations with high uncertainty that could not have been observed previously.
By utilizing problems with known underlying target and noise functions, which are unknown to the AS methods, we can simulate and evaluate accurately the performance improvements resulting from the decisions made by each method in previous iterations.

\subsection{Experiments with One-Dimensional Data} \label{sec:1d}

\begin{figure}[t]
    \centering
    \includegraphics[width=\columnwidth]{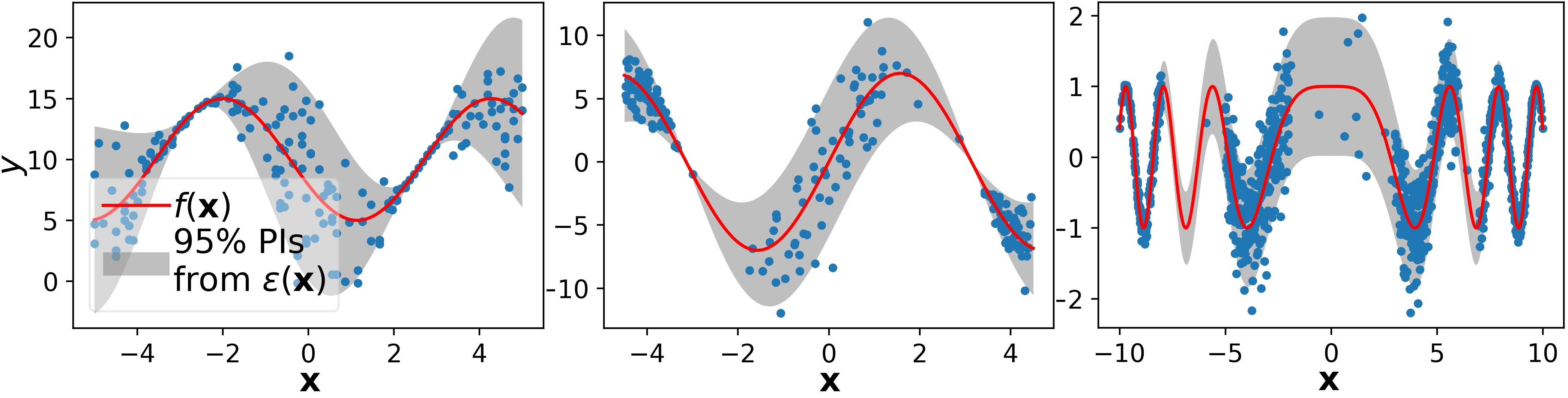}
    \caption{Initial \texttt{cos}, \texttt{hetero}, and \texttt{cosqr} datasets and the ideal 95\% PIs calculated from $\varepsilon_a(\mathbf{x})$ across the domain.}
    \label{fig:datasets}
\end{figure}

We considered three 1-D problems: \texttt{cos}~\cite{DualAQD}, \texttt{hetero}~\cite{depeweg18a}, and \texttt{cosqr}.
All three problems are affected by heteroscedastic noise, and their function equations are shown in Table~\ref{tab:1Dproblems}.
Unlike most AL and AS approaches, we do not initiate the experiments from empty datasets.
For each case, we generated incomplete datasets as initial states, as shown in Fig.~\ref{fig:datasets}.
The motivation for this is to produce areas with low data density, which entails high epistemic uncertainty.
Thus, methods that estimate potential epistemic uncertainty more accurately and select sampling locations designed to reduce such uncertainty should require fewer AS iterations to approximate the ground-truth distribution of the problem.
Additional implementation details are provided in the Appendix.


\begin{table}[t]
\small
\centering
\caption{Functions and noise terms of the 1-D problems.}
    \label{tab:1Dproblems}
\def\arraystretch{1.2}
\begin{tabular}{|c|c|c|}
\hline
\textbf{Name} & \textbf{Function} $f(\mathbf{x})$ & \textbf{Noise} $\varepsilon_a(\mathbf{x})$ \\
\hline
\texttt{cos} & $10 + 5 \cos(\mathbf{x} + 2)$ & $\mathcal{N}(0,2 + 2 \cos(1.2 \mathbf{x}))$ \\
\hline
\texttt{hetero} & $7\sin(\mathbf{x})$ & $\mathcal{N}(0,3 \cos(\mathbf{x}/2))$ \\
\hline
\texttt{cosqr} & $10 + 5 \cos(\frac{\mathbf{x}^2}{5})$ & $\mathcal{N}(0,\frac{1}{2} (1 - \frac{\mathbf{x}^2}{100}))$ \\
\hline
\end{tabular}
\end{table}

For ASPINN, we trained feed-forward NNs with varying depths: two hidden layers with 100 units for problems \texttt{cos} and \texttt{hetero}; and three hidden layers with 500, 100, and 50 units, respectively, for \texttt{cosqr}.
The networks $\hat{f}_t$ and $\hat{g}_t$ share the same architecture except for the last layer, as $\hat{f}_t$ uses one output, while $\hat{g}_t$ uses two outputs.
Furthermore, ASPINN uses two hyperparameters: the neighbor distance threshold $\theta$ and the kernel length $r$.
We performed a grid search with the values $\theta=[0.1, 0.15, 0.2, 0.25]$ and $r=[0.1, 0.15, 0.2, 0.25]$, and selected $\theta=0.25$ and $r=0.15$ for all experiments.
DualAQD, the PI-generation method used by ASPINN, uses a hyperparameter $\eta$ as a scale factor to adapt the coefficient that balances the two objectives of the DualAQD loss function.
We chose a scale factor $\eta=0.1$.
Other $\eta$ values (i.e., $\{0.001, 0.005, 0.01, 0.05, 0.1\}$) achieved similar results but with slower convergence rates.

For MC-Dropout, we used the same architecture as the target-estimation NN in ASPINN.
For NF-Ensemble, we used flows with 200 hidden units for problems 
\texttt{cos} and \texttt{hetero}
and 300 hidden units for problem \texttt{cosqr}.
We employed ensembles consisting of five models trained during 30,000 epochs.
For the standard GP, we used the same RBF kernel used by ASPINN.
We utilized an inference implementation based on black-box matrix-matrix multiplication~\cite{Gpytorch} that uses 3000 training epochs.

Our objective is to reduce the epistemic uncertainty with as few AS iterations as possible.
We define the performance metric $PI^{(t)}_\delta$ to quantify epistemic uncertainty relative to the ground truth at the $t$-th iteration: 
\[
PI^{(t)}_\delta = \frac{1}{\left| \mathcal{X} \right|} \sum_{\mathbf{x} \in \mathcal{X}} \left( |y^u(\mathbf{x}) - \hat{y}_t^u(\mathbf{x})| + | y^\ell(\mathbf{x}) - \hat{y}_t^\ell(\mathbf{x})| \right).
\]
Here, $y^\ell(\mathbf{x})$ and $y^u(\mathbf{x})$ represent the ideal lower and upper PI bounds, respectively, calculated from the aleatoric noise function: $y^u(\mathbf{x}) = f(\mathbf{x}) + 1.96 \, \varepsilon_a(\mathbf{x})$ and $y^\ell(\mathbf{x}) = f(\mathbf{x}) - 1.96 \, \varepsilon_a(\mathbf{x})$.
This metric is applicable to problems with normally distributed aleatoric noise, which is the case for the problems evaluated in this work. 
However, none of the tested methods make assumptions about the noise distribution.
Note that if $PI^{(t)}_\delta = 0$, the estimated PIs match the ideal intervals, implying that the model's epistemic uncertainty has been minimized, and the total uncertainty is purely aleatoric.
A non-zero $PI^{(t)}_\delta$ indicates a discrepancy between the estimated and ideal PIs, suggesting the presence of epistemic uncertainty.
The greater the $PI^{(t)}_\delta$, the higher the epistemic uncertainty.
To ensure fairness, $\hat{y}_t^\ell(\mathbf{x})$ and $\hat{y}_t^u(\mathbf{x})$ are generated by an independent NN, $\hat{g}_t(\cdot)$, trained on the dataset $\mathcal{D}_t$ produced by each compared method at each iteration.
Regardless of the uncertainty estimation model used by each method, we trained an additional PI-generation NN using the DualAQD loss to maintain a consistent uncertainty metric across all comparisons.

It is worth mentioning that other works have used different evaluation approaches. 
For instance, \citet{NFs} employed an approach where they sampled 50 random locations from the domain.
For each location, they generated 1000 samples using the ground-truth distribution and 1000 samples using the distribution predicted by each method.
They then calculated the Kullback-Leibler divergence between the ground truth and the model-generated distributions.
However, we believe this approach does not provide a consistent basis for evaluation, as each method employs different mechanisms for estimating uncertainty.

\begin{figure}[t]
    \centering
    \includegraphics[width=\columnwidth]{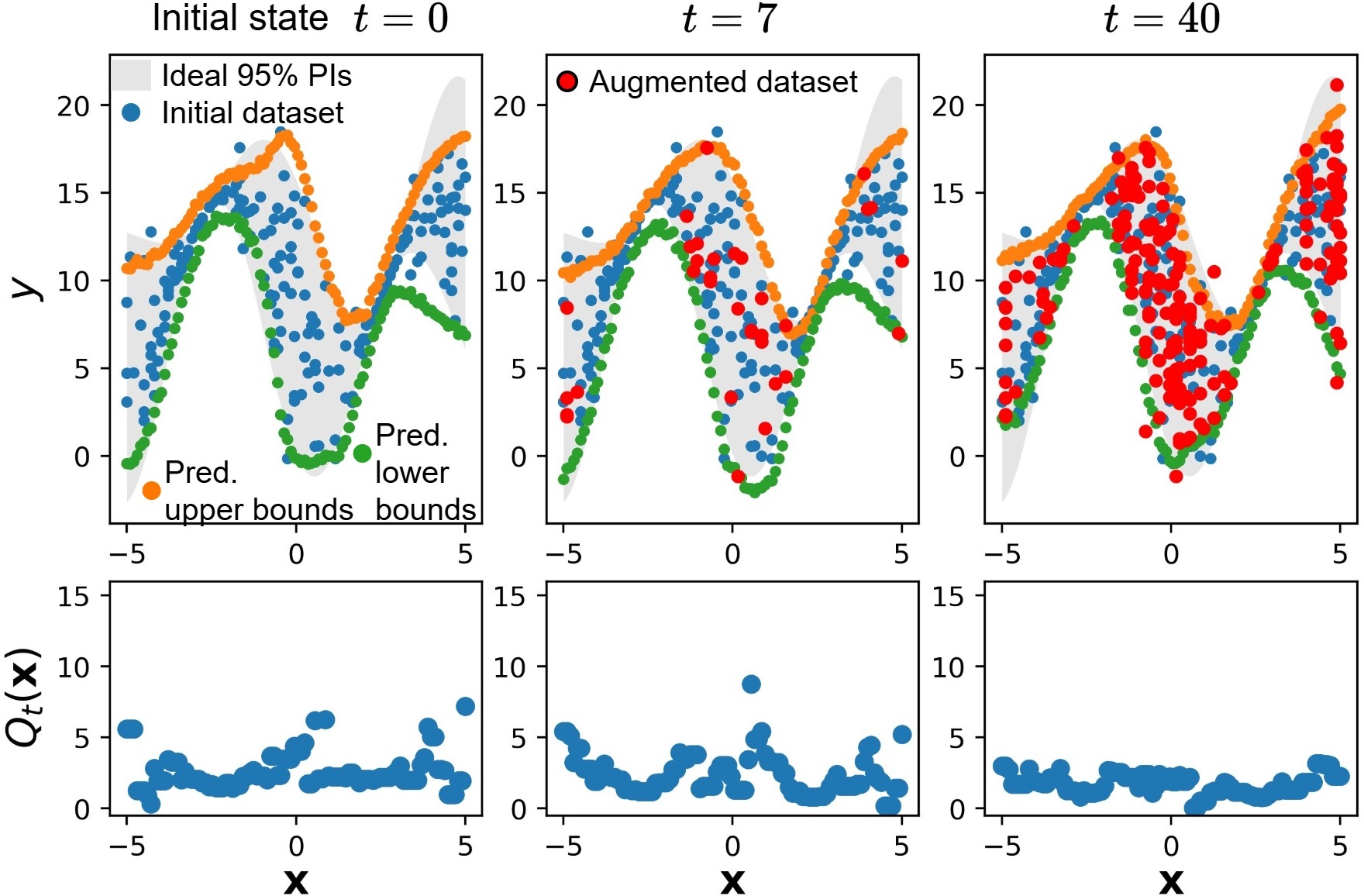}
    \caption{Example of the adaptive sampling process using ASPINN on the \texttt{cos} problem.}
    \label{fig:ASPINN}
\end{figure}

\begin{figure}[t]
    \centering
    \includegraphics[width=0.85\columnwidth]{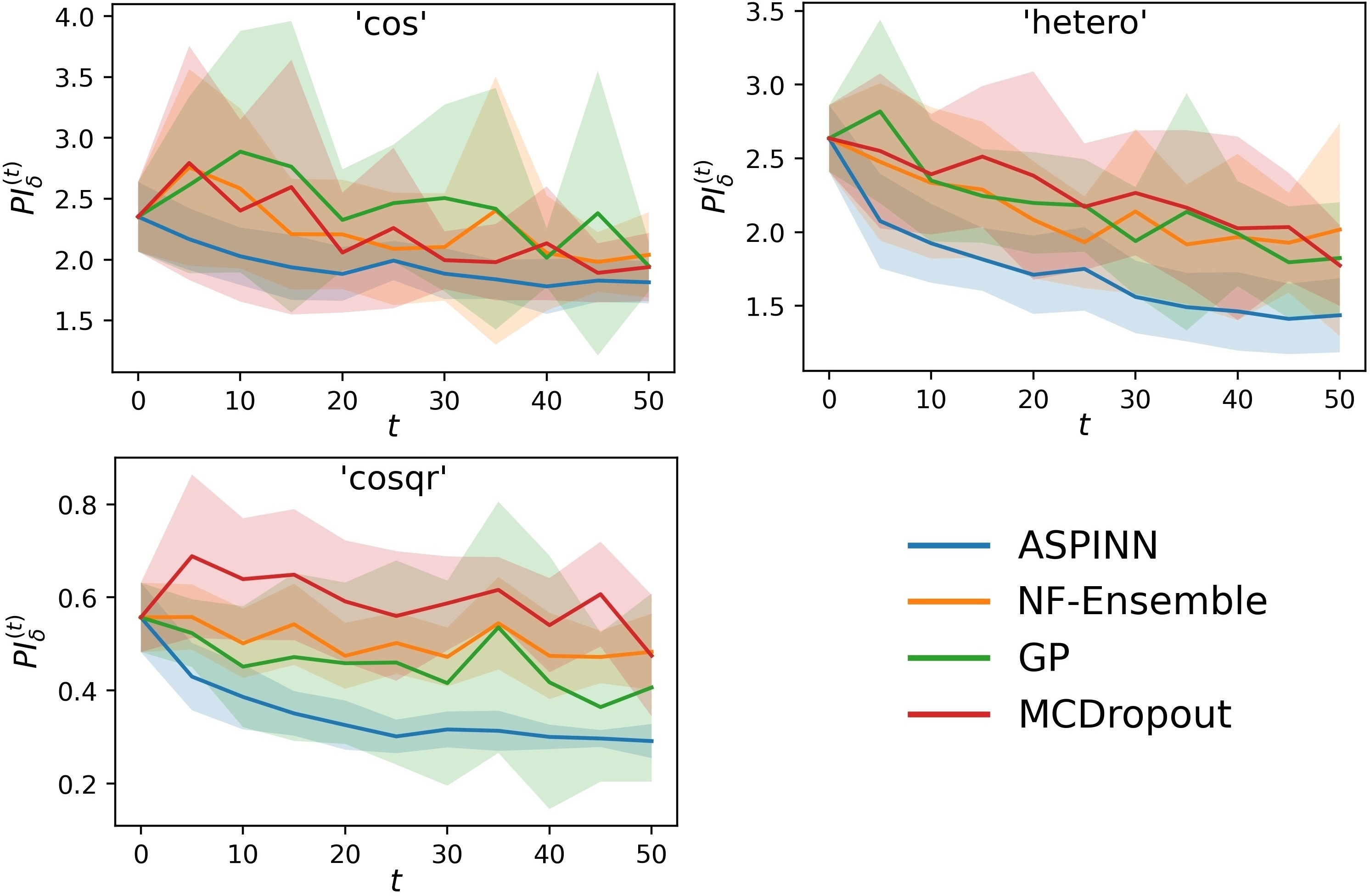}
    \caption{Evolution of the mean $PI^{(t)}_\delta$ value and its corresponding standard deviation for the 1-D problems.}
    \label{fig:results}
\end{figure}

For our experiments, the AS process was executed for each problem for 50 iterations. 
This process is repeated 10 times, initializing the problems with a different seed each time.
Figure~\ref{fig:ASPINN} depicts an initial state of problem \texttt{cos} along with the augmented datasets during iterations $t=7,\,40$. 
The figure also displays the corresponding calculated potential epistemic uncertainty for all values of the input domain.
Figure~\ref{fig:results} shows the evolution of the mean $PI^{(t)}_\delta$ value and its corresponding standard deviation, calculated across the values obtained from the 10 repetitions at each $t$. 
In addition, we calculated the area under the uncertainty curve (AUUC) for each learning curve.
For each problem, Table~\ref{tab:1D} gives the average AUUC for the four methods and corresponding standard deviations.
The bold entries indicate the method that achieved the lowest average AUUC value and that its difference with respect to the values obtained by the other methods is statistically significant according to a paired $t$-test performed at the 0.05 significance level.

\begin{table}[t]
    \centering
    \small 
    \caption{AUUC comparison for the 1-D problems}
    \label{tab:1D}
    \resizebox{\columnwidth}{!}{
    \def\arraystretch{1.}
    \begin{tabular}{|c|c|c|c|c|}
        \hline
        \textbf{Problem} & \textbf{MCDropout} &\textbf{GP} & \textbf{NF-Ensemble} & \textbf{ASPINN} \\ \hline
        \texttt{cos} & 112.57$\pm$24.20 & 123.87$\pm$26.22 & 113.39$\pm$19.49 & \textbf{97.26$\pm$7.87} \\ \hline
        \texttt{hetero} & 113.80$\pm$13.38 & 110.21$\pm$13.59 & 106.44$\pm$16.26 & \textbf{85.95$\pm$9.11} \\ \hline
        \texttt{cosqr} & 30.39$\pm$3.56 & 23.12$\pm$5.55 & 25.60$\pm$2.67 & \textbf{17.13$\pm$1.42} \\ \hline
        \end{tabular}%
    }
\end{table}

\subsection{Experiments with Simulated Field Data} \label{sec:MZ}

In this section, we present a multi-dimensional problem that simulates a real-world agricultural field site.
A field site is defined as a specific area within a larger field (e.g., a $10\times10$\SI{}{\meter}). 
It is used for precise monitoring and management to address local variations in soil and crop conditions.

Note that actual real-world data cannot be considered for a comparative AS study.
There are multiple reasons for this.
First, a given field site receives a single experimental rate during the fertilization stage and its effects are observed during the harvest season (e.g., five months for winter wheat).  
Second, additional samples at the same site require collecting data over multiple years. 
Third, when comparing different AS methods, they may produce different experimental rates, which cannot be implemented simultaneously in a single season.
Fourth, real-world conditions, such as unforeseen environmental factors and concept drift, introduce additional complexity, making it difficult to isolate the AS strategies' effects. 
Therefore, simulations based on the properties of a real field provide a controlled environment where different AS methods can be evaluated under identical conditions, allowing for a fair comparison.

In previous work, we derived the functional form of N-response curves of different management zones (MZs) from an actual winter wheat field as symbolic skeleton expressions using a Multi-Set Transformer~\cite{MultiSetSR}.
An MZ is defined as a distinct sub-region that encompasses sites with relative homogeneity and, thus, similar fertilizer responsivity (i.e., similar response to varying fertilizer rates).
A symbolic skeleton expression is a representation of a mathematical expression that captures its structural form without setting specific numerical values.
For instance, the relationship between yield, $y$, and N rate, $\mathbf{x}^{Nr}$, at a given site is given by the skeleton $y=c_1 + c_2\, \texttt{tanh}(c_3 + c_4\, \mathbf{x}^{Nr})$, where $c_1$--$c_4$ are placeholder constants.
In this work, we propose to use a simulated field site from an MZ whose underlying function is based on the previous skeleton.

In particular, we consider the following yield function:
\[
y=f(\mathbf{x}) = \frac{\mathbf{x}^{P}}{15} + \left(\frac{\mathbf{x}^{A}}{\pi} + 1\right) \tanh\left(\frac{0.1 \, \mathbf{x}^{Nr}}{3 \mathbf{x}^{VH} + 2}\right) + \varepsilon_a(\mathbf{x}),
\]
where $\mathbf{x} = [\mathbf{x}^{P}, \mathbf{x}^{A}, \mathbf{x}^{VH}, \mathbf{x}^{Nr}]$ comprises the following site-specific covariates: annual precipitation (mm), terrain aspect (radians), Sentinel-1 backscattering coefficient from the Vertical Transmit-Horizontal Receive Polarization band, and applied N rate (lbs/ac), respectively.
The aleatoric noise is modeled as $\varepsilon_a(\mathbf{x}) = \mathcal{N}(0, (\mathbf{x}^{P} + \mathbf{x}^{Nr}) / 150)$.
Further details on the selection of these underlying and noise functions are available in the Appendix.

While this yield regression problem considers four explanatory variables, the only 
one that farmers can control is $\mathbf{x}^{Nr}$.
Therefore, the AS search is focused along the $\mathbf{x}^{Nr}$ axis to determine the best experimental N rate for reducing epistemic uncertainty.
A field site receives a single fertilizer treatment; thus, we consider $B=1$.
The AS process was conducted over 50 iterations, with each iteration representing a different year or growing season, corresponding to a randomly generated precipitation value $\mathbf{x}^{P} \sim U(75, 150)$.
All compared methods used the same sequence of precipitation values throughout the AS process.
$\mathbf{x}^{A}$ describes topographic information of the field so it is assumed to remain constant throughout all iterations.
In contrast, $\mathbf{x}^{VH}$, associated with soil moisture, was modeled as a function of precipitation and topographic aspect. 
Additional details on data generation are provided in the Appendix.

\begin{figure}[t]
    \centering
    \includegraphics[width=0.6\columnwidth]{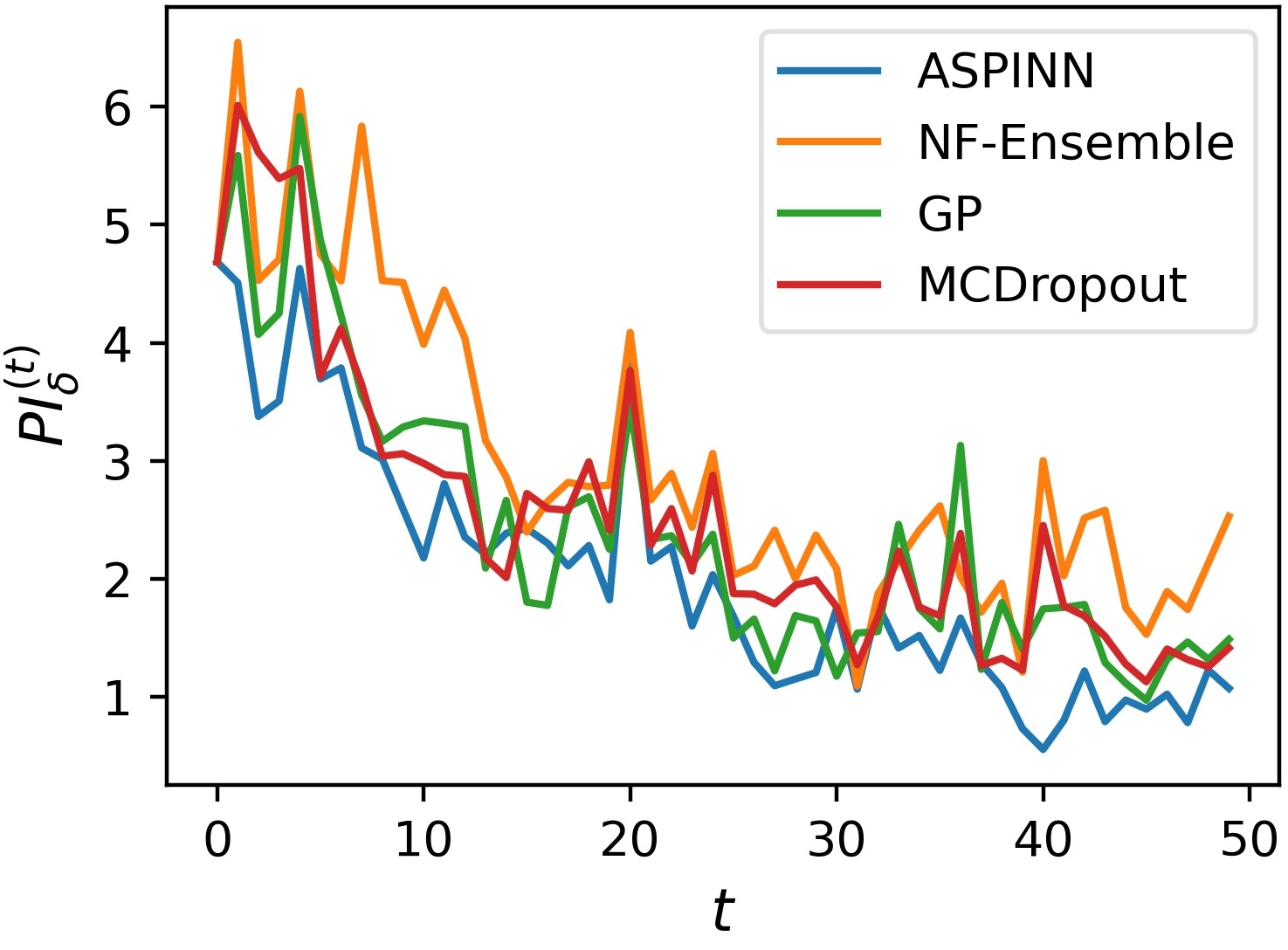}
    \caption{Evolution of the mean $PI^{(t)}_\delta$ value and its corresponding standard deviation for the simulated field site.}
    \label{fig:resultsMZ}
\end{figure}

We applied the AS process ten times.
At each iteration, we used a unique initialization seed and evaluated the epistemic uncertainty along the allowed N rates (i.e., 0, 30, 60, 90, 120, and 150 lbs/ac) under the current field conditions.
Table~\ref{tab:MZ} presents the average AUUC values and corresponding standard deviations, highlighting the best-performing method in bold.
Figure~\ref{fig:resultsMZ} depicts the evolution of the mean $PI^{(t)}_\delta$ values, calculated based on the results from the ten repetitions. 

\begin{table}[t]
    \centering
    \small 
    \caption{AUUC comparison for the simulated field site}
    \label{tab:MZ}
    \resizebox{\columnwidth}{!}{
    \def\arraystretch{1.1}
    \begin{tabular}{|c|c|c|c|}
        \hline
        \textbf{MCDropout} &\textbf{GP} & \textbf{NF-Ensemble} & \textbf{ASPINN} \\ \hline
        $614.68\pm112.48$ & $593.54\pm107.42$ & $730.80\pm74.63$ & \textbf{496.85$\pm$71.65} \\ \hline
        \end{tabular}%
    }
\end{table}

\section{Discussion}

The ASPINN method involves training a PI-generation NN, which is used to design a novel potential epistemic uncertainty metric.
This metric is then used in our batch sampling strategy to determine the sequence of sampling locations most likely to reduce epistemic uncertainty the greatest across the input domain.

When evaluating ASPINN on the tested 1-D problems, as shown in Fig.~\ref{fig:results}, we observed that it produced learning curves with faster convergence rates and lower standard deviation than the other methods.
Although the confidence bands exhibit some overlap, this is attributed to outliers with high $PI^{(t)}_\delta$ values generated by other methods (e.g., GP), which increase the variance.
Nevertheless, it is important to note that the learning curves for ASPINN consistently remain below those of the other methods across all iterations and have narrower confidence bands.
Thus, the difference in AUUC values is shown to be statistically significant according to the $t$-test, as shown in Table~\ref{tab:1D}.
Also from Fig.~\ref{fig:results}, we notice that ASPINN generated constantly decreasing and smoother learning curves.
Conversely, other methods, such as MC-Dropout, tend to oversample certain regions of the input domain, leading to imbalanced datasets.
This oversampling results in overfitting in those regions while causing a poor fit in others, producing unstable learning curves.

Furthermore, the experiments conducted on the simulated field data exhibit consistent behavior with the results from the 1-D problems
In particular, Table~\ref{tab:MZ} demonstrates that ASPINN achieves the lowest AUUC values, and the differences between ASPINN and the compared methods are statistically significant.
Given that the precipitation values vary at each iteration, the resulting learning curves are expected to exhibit multiple peaks and valleys rather than a smooth, consistently decreasing trend, as observed in Fig~\ref{fig:resultsMZ}.
This variability arises because higher precipitation values are associated with increased uncertainty levels, leading to more pronounced fluctuations in the learning curves.
Considering that the sequence of precipitation values is not the same for all AS repetitions, Fig~\ref{fig:resultsMZ} reports only the mean curve and not the confidence bands.
This is because the $PI^{(t)}_\delta$ values obtained by a method across different iterations are generated from contexts that could correspond to extreme opposites, leading to high variance values that do not necessarily reflect the method's performance.
Despite this behavior, we observed that ASPINN consistently produced learning curves that remained below those of the compared methods.

One limitation of our approach is that it does not handle multi-modal aleatoric noise inherently.
Multi-modal noise indicates that the data variability comes from different underlying sources, each contributing to a different mode in the noise distribution.
In such cases, it would be necessary to use a PI-generation method capable of producing multiple upper and lower bounds based on the identified number of modes.
Note, however, that the contributions proposed in this paper are not reliant on a specific PI-generation method.
In the presence of multiple PIs, we would need to adapt the epistemic uncertainty metric accordingly and execute the remaining steps similarly.
Another limitation, which also applies to the compared methods, is the computational cost when dealing with high-dimensional problems due to the need to evaluate all potential locations in the input space. 
We plan to address this limitation in future work.

\section{Conclusion}

Accurate predictive modeling is essential in many scientific and engineering disciplines, where decisions often rely on data gathered from costly and time-consuming experiments.
This is especially true in fields like precision agriculture, where data collection is limited by factors such as growing seasons and crop rotation.
In such contexts, reducing uncertainty in prediction models is necessary for optimizing outcomes and ensuring reliable decision-making.
Addressing this challenge, our work focuses on minimizing epistemic uncertainty through adaptive sampling techniques.

We introduced ASPINN, an adaptive sampling technique designed to reduce epistemic uncertainty across an input domain using prediction intervals generated by neural networks.
The novel potential epistemic uncertainty metric, central to ASPINN, provided a robust basis for guiding the sampling process. 
The effectiveness of our approach was demonstrated through its consistent ability to achieve faster convergence rates with lower and more stable learning curves compared to other methods.
This was observed across all tested scenarios, including 1-D synthetic problems and a multi-dimensional problem that simulates an agricultural field site based on real-world winter wheat data.

In the future, we plan on adapting ASPINN for problems affected by both heteroskedastic and multi-modal noise.
In particular, this would involve integrating PI-generation techniques capable of addressing multi-modal noise functions and refining the potential epistemic uncertainty metric to account for multiple PIs at a single location.

\section{Acknowledgments}
This research was supported by the Data Intensive Farm Management project (USDA-NIFA-AFRI 2016-68004-24769 and USDA-NRCS NR213A7500013G021).
Computational efforts were performed on the Tempest HPC System, operated by University Information Technology Research Cyberinfrastructure at MSU.

\bibliography{aaai25}

\newpage
\appendix

\renewcommand{\thesection}{A}
\counterwithin{figure}{section}
\counterwithin{table}{section}
\counterwithin{algorithm}{section}
\counterwithin{equation}{section}

\section*{\huge Appendix}

\vspace{.5ex}

In this supplementary material, we provide pseudocode for the main functions of our method, Adaptive Sampling with Prediction-Interval Neural Networks (ASPINN).
In addition, we present experimental details for the synthetic 1-D and simulated field site problems presented in the paper.

\subsection{ASPINN Algorithms}

Algorithm~\ref{alg:epist} describes the function \texttt{PotEpistUnc} that calculates the potential epistemic uncertainty at each candidate position for sampling.
It takes as inputs the dataset $\mathcal{D}_t = (\mathbf{X}_{obs}^{(t)}, \mathbf{Y}_{obs}^{(t)})$ available during the $t$-th iteration of the adaptive sampling (AS) process, the prediction interval (PI)-generation neural network (NN) $g_t(\cdot)$ trained on $\mathcal{D}_t$, the set $X_{test}$ of all candidate positions for sampling (i.e., the input space), and the neighbor distance threshold $\theta$. 

For each candidate position $\mathbf{x}_p$, the algorithm constructs a neighborhood $\mathcal{N}(\mathbf{x}_p)$ consisting of all samples in $\mathbf{X}_{obs}^{(t)}$ within a radius of $\theta$ with respect to $\mathbf{x}_p$.
If  $\mathcal{N}(\mathbf{x}_p)$ is empty, the potential epistemic uncertainty is given by the PI width $\hat{y}_t^u(\mathbf{x}_p) - \hat{y}_t^\ell(\mathbf{x}_p)$.  
Otherwise, it builds the set of input--response pairs $\mathcal{R}(\mathcal{N}(\mathbf{x}_p))$ using the samples in $\mathcal{N}(\mathbf{x}_p)$ whose response values fall within their corresponding PI.
From the data points in $\mathcal{R}(\mathcal{N}(\mathbf{x}_p))$, the potential epistemic uncertainty $Q_t(\mathbf{x}_p)$ is calculated as the sum of the minimum distance between the predicted upper bounds and the observed values, and the minimum distance between the observed values and the predicted lower bounds.

Furthermore, ASPINN's batch sampling strategy is shown in Algorithm~\ref{alg:sample}.
It takes as inputs the set $X_{test}$ of all candidate positions for sampling, their corresponding potential epistemic uncertainty values $Q_t$ during the $t$-th iteration of the AS process, the batch size $B$, and the kernel length $r$.
In Lines~\ref{line:init1}--\ref{line:init2}, the algorithm initializes the covariance matrix $\mathbf{K}_t$ of a Gaussian Process (GP) surrogate model. 
The diagonal of $\mathbf{K}_t$ is set to be equal to $Q_t$. 
Since the uncertainty at a given position may be correlated with nearby positions, $\mathbf{K}_t$ is structured as a matrix with non-zero off-diagonal elements.  
The off-diagonal elements combine the potential epistemic uncertainty values at different positions based on their correlation value, which is calculated using a radial base function (RBF) with kernel length $r$.

Once $\mathbf{K}_t$ is initialized, we assess the potential uncertainty reduction when observing each candidate position $\mathbf{x}_p$.
Specifically, we condition the GP on a data point at $\mathbf{x}_p$ and update its covariance matrix as shown in Line~\ref{line:K}.
The total potential epistemic uncertainty across the domain, $J(\mathcal{D}_t)$, is determined by summing the diagonal elements $\mathbf{K}_t$.
The estimated reduction in the total potential epistemic uncertainty, when making an observation at $\mathbf{x}_p$, is thus calculated as the difference $\delta J$ between the sum of the diagonal elements of $\mathbf{K}_t$ before and after the observation (Line~\ref{line:deltaJ}).
Therefore, the $k$-th element of the $t$-th batch, $\mathbf{x}_{t, k}$, is selected as the position that yields the greatest $\delta J$ value.

\begin{algorithm} [!t]
\small
\renewcommand{\baselinestretch}{1.2}\selectfont
\caption{ASPINN's potential epistemic uncertainty }
\begin{algorithmic}[1]
\Function{PotEpistUnc}{$\mathcal{D}_t, g_t, X_{test}, \theta$}
    \State $\left(\mathbf{X}_{obs}^{(t)}, \mathbf{Y}_{obs}^{(t)}\right) \leftarrow \mathcal{D}_t$
    \State $Q_t \leftarrow \text{zeros}(\text{size}(X_{test}))$
    \For{$\mathbf{x}_p \in X_{test}$}
        \State $\mathcal{N}(\mathbf{x}_p) \leftarrow \{ \mathbf{x} \in \mathbf{X}_{obs}^{(t)} | \, \| \mathbf{x} - \mathbf{x}_p \|_2 \leq \theta\}$ 
        \If{$\mathcal{N}(\mathbf{x}_p) \neq \emptyset$}
            \State $\mathcal{R}(\mathcal{N}(\mathbf{x}_p)) \leftarrow [\,]$ \Comment{\scriptsize $\mathbf{x}_p$'s neighbors falling within the PIs \small}
            \State $Y_{\text{sub}}^u, Y_{\text{sub}}^{\ell} \leftarrow [\,], [\,]$
            \For{$\mathbf{x} \in \mathcal{N}(\mathbf{x}_p)$}
                \State $\hat{y}^\ell(\mathbf{x}), \hat{y}^u(\mathbf{x}) \leftarrow g_t (\mathbf{x})$
                \If{$\hat{y}^\ell(\mathbf{x}) \leq y \leq \hat{y}^u(\mathbf{x})$} \Comment{\scriptsize $(\mathbf{x}, y) \in \mathcal{D}_t$ \small}     
                    \State $\mathcal{R}(\mathcal{N}(\mathbf{x}_p))$.append$\left( (\mathbf{x}, y )\right)$
                    \State $Y_{\text{sub}}^u$.append$\left( \hat{y}^u(\mathbf{x})\right)$
                    \State $Y_{\text{sub}}^{\ell}$.append$\left( \hat{y}^{\ell}(\mathbf{x})\right)$
                \EndIf
            \EndFor
            \State $(X_{\text{sub}} \, , Y_{\text{sub}}) \leftarrow \mathcal{R}(\mathcal{N}(\mathbf{x}_p))$
            \State $Q_t(\mathbf{x}_p) \leftarrow \min(Y_{\text{sub}}^u - Y_{\text{sub}}) + \min(Y_{\text{sub}} - Y_{\text{sub}}^{\ell})$
        \Else
            \State $\hat{y}^\ell(\mathbf{x}_p), \hat{y}^u(\mathbf{x}_p) \leftarrow g_t (\mathbf{x}_p)$
            \State $Q_t(\mathbf{x}_p) \leftarrow \hat{y}_t^u(\mathbf{x}_p) - \hat{y}_t^\ell(\mathbf{x}_p)$
        \EndIf
    \EndFor
    \State \Return $Q_t$
\EndFunction
\end{algorithmic}
\label{alg:epist}
\end{algorithm}

\begin{algorithm} [t]
\small
\renewcommand{\baselinestretch}{1.1}\selectfont
\caption{ASPINN's batch sampling method}
\begin{algorithmic}[1]
\Function{sample}{$X_{test},  Q_t, B, r$}
    \State $n_Q \leftarrow \text{size}(Q_t)$
    \State $\mathbf{K}_t \leftarrow \text{zeros}(n_Q, n_Q)$ \Comment{\scriptsize Init GP's covariance matrix \small}
    \label{line:init1}
    \For{$i \in (0, n_Q)$}
        \For{$j \in (i, n_Q)$}
            \If{$i = j$}
                \State $k \leftarrow Q_t(X_{test}[i])$
            \Else
                \State $\rho \leftarrow \text{RBF}(X_{test}[i], X_{test}[j];r)$ \Comment{\scriptsize $r$: kernel size \small}
                \State $k \leftarrow \rho\sqrt{Q_t(X_{test}[i]) Q_t(X_{test}[j])}$
            \EndIf
            \State $\mathbf{K}_t(i, j) = \mathbf{K}_t(k, i) = k$
        \EndFor
    \EndFor
    \label{line:init2}
    
    \State $\mathbf{X}_{acq}^{(t)} \leftarrow [\,]$

    \While{$\text{size}(\mathbf{X}_{acq}^{(t)}) < B$}  \Comment{\scriptsize Batch sampling loop \small}
        \State $\Delta J_{\text{max}} \leftarrow [\,]$
        \For{$\mathbf{x}_p \in X_{test}$}
            \State $\mathbf{K}'_t \!\leftarrow\! \mathbf{K}_t(\mathbf{x}, \mathbf{x}') \!-\! \mathbf{K}_t(\mathbf{x}, \mathbf{x}_p) \mathbf{K}_t(\mathbf{x}_p, \mathbf{x}_p)^{-1}$
            \label{line:K}
            \State \hspace{2.3em} $\mathbf{K}_t(\mathbf{x}_p, \mathbf{x}')$ \Comment{\scriptsize $\forall \mathbf{K}_t(\mathbf{x}, \mathbf{x}') \in \mathbf{K}_t$ \small}
            \State $\Delta J \leftarrow \sum{\text{diag}(\mathbf{K})} - \sum{\text{diag}(\mathbf{K}')}$
            \label{line:deltaJ}
            \If{$\Delta J > \Delta J_{\text{max}}$}
                \State $\Delta J_{\text{max}} \leftarrow \Delta J$
                \State $\mathbf{x}_{t, k} \leftarrow \mathbf{x}_p$
                \State $\mathbf{K}_{\text{best}} \leftarrow \mathbf{K}'_t$
            \EndIf
        \EndFor
        \State $\mathbf{X}_{acq}^{(t)}$.append($\mathbf{x}_{t, k}$)
        \State $\mathbf{K}_t \leftarrow \mathbf{K}'_t$
    \EndWhile
    
    \State \Return $\mathbf{X}_{acq}^{(t)}$
\EndFunction
\end{algorithmic}
\label{alg:sample}
\end{algorithm}

\subsection{Experiments with One-Dimensional Data} \label{sec:1dap}

In this section, we first provide details on the generation of the 1-D test problems. 
We also present additional experimental results and corresponding plots.

\subsubsection{Data Generation}

Below, we report the admissible search space for each problem:

\begin{itemize}
    \item \texttt{cos}: $\mathcal{X} = \left\{ -5 + \frac{10(i-1)}{99}, \mid i = 1, 2, \dots, 100 \right\}$
    \item \texttt{hetero}: $\mathcal{X} = \left\{ -4.5 + \frac{9(i-1)}{299}, \mid i = 1,2, \dots, 300 \right\}$
    \item \texttt{cosqr}: $\mathcal{X} = \left\{ -10 + \frac{20(i-1)}{499}, \mid i = 1,2, \dots, 500 \right\}$
\end{itemize}

For the case of the \texttt{cos} problem, we generate the initial set of observations $\mathbf{X}_{obs}^{(t=0)}$ by uniformly sampling 200 elements from the discrete set $\mathcal{X}$.

The initial datasets corresponding to the \texttt{hetero} problem are generated as recommended by \citet{depeweg18a}.
In particular, a mixture of three Gaussian is created with means $\mu_1 = -4$, $\mu_2 = 0$, and $\mu_3 = 4$ and corresponding variances $\sigma_1 = \frac{2}{5}$, $\sigma_2 = 0.9$, and $\sigma_1 = \frac{2}{5}$.
Each Gaussian component is equally weighted.
We considered an initial dataset size of $|\mathbf{X}_{obs}^{(t=0)}|=200$. 

For the \texttt{cosqr} problem, the initial dataset is generated by first sampling 2,000 elements from the discrete set $\mathcal{X}$ uniformly and then applying a series of masks to select specific ranges of values.
The process is as follows:

\begin{itemize}
\item Elements in the intervals $[-10, -8)$, $[-5, -2)$, $[3, 6)$, and $[7, 10]$ are included in the dataset directly.
\item Additional elements are selected from the intervals $[-8, -5)$, $[-2, 3)$, and $[6, 7)$ with specific sizes of 1, 10, and 3 elements, respectively.
\end{itemize}

\begin{figure}
    \centering
    \includegraphics[width=\linewidth]{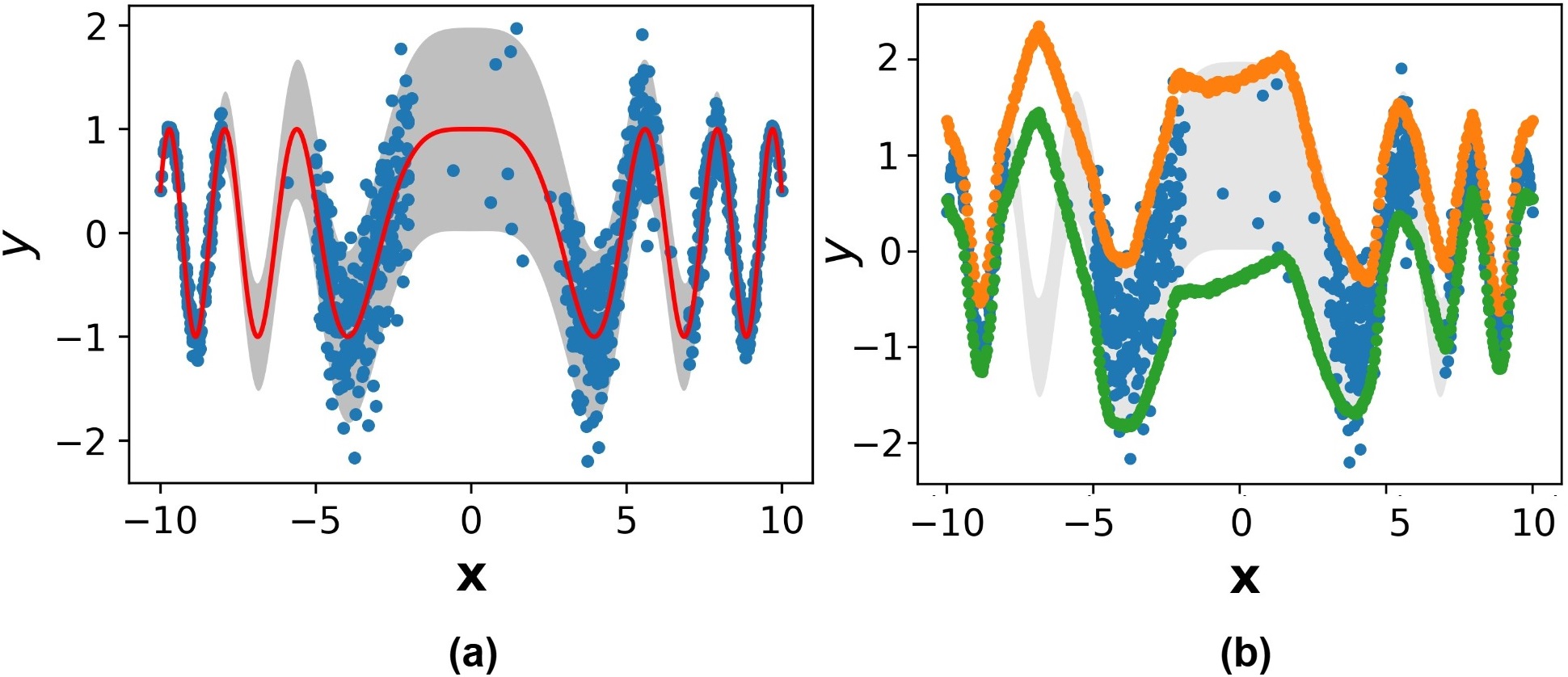}
    \caption{\texttt{cosqr} problem. \textbf{(a)} An initial generated dataset and the ideal 95\% PIs calculated from $\varepsilon_a(\mathbf{x})$ across the domain. \textbf{(b)} Initial PIs estimated using DualAQD.}
    \label{fig:cosqr}
\end{figure}

By doing so, we aim to generate a complex dataset with different areas with low data density, as depicted in Fig~\ref{fig:cosqr}.
Note that the low-density regions correspond to distinct behaviors in both the function $f(\mathbf{x})$ and the noise $\varepsilon_a(\mathbf{x})$.
This approach ensures that the AS process remains focused on capturing meaningful variations in the data, rather than merely estimating data density for selecting future sample locations.
For example, the intervals $[-8, -5)$ and $[6, 7)$ each contain only a single observed point. 
However, the former covers an entire oscillation of the function, whereas the latter spans a much smaller range. 
Consequently, when using a PI-generation neural network to analyze the unobserved areas, there is a greater discrepancy between the estimated and ideal PIs in the first case.
All PI-generation NNs are trained using a specialized loss function called DualAQD~\cite{DualAQD}.

Furthermore, the initial dataset size for problem \texttt{cosqr} varies according to the selected initialization seed.
Specifically, the obtained sizes $|\mathbf{X}_{obs}^{(t=0)}|$ for the ten AS iterations are: 1,102, 1,106, 1,123, 1,078, 1,114, 1,163, 1,084, 1,159, 1,079, and 1,104, respectively.

\subsubsection{Experimental Results}

In the paper, we reported that ASPINN achieved the lowest average area under the uncertainty curve (AUUC) for the learning curves obtained for each problem.
In addition, its difference with respect to the values obtained by the compared methods was found to be statistically significant according to a paired $t$-test performed at the 0.05 significance level.
The compared methods are Normalizing flows ensembles (NF-Ensemble)~\cite{NFs}, a standard GP~\cite{Gpytorch}, and MC-Dropout~\cite{pmlr-v48-gal16}.

To demonstrate these results, Table~\ref{tab:1Dstats} presents the $p$-values from the paired $t$-tests comparing ASPINN with the other methods.
Here, the upward-pointing arrow ($\uparrow$) indicates that ASPINN performed significantly better (i.e., $p$-value $<0.05$).

\begin{table}[t]
\centering
\caption{Statistical significance tests --- 1-D problems. $p$-values obtained comparing ASPINN to the other methods.}
\label{tab:1Dstats}
\def\arraystretch{1.3}
\begin{tabular}{|c|c|c|c|}
\hline
\begin{tabular}[c]{@{}c@{}}\textbf{Compared}\\ \textbf{Method}\end{tabular} & \texttt{cos} & \texttt{hetero} & \texttt{cosqr} \\ \hline
NF-Ensemble & 4.4E-2 ($\uparrow$) & 6.4E-3 ($\uparrow$) & 5.3E-6 ($\uparrow$) \\ \hline
GP & 5.4E-3 ($\uparrow$) & 8.6E-4 ($\uparrow$) & 7.7E-3 ($\uparrow$) \\ \hline
MC-Dropout & 3.8E-2 ($\uparrow$) & 5.7E-4 ($\uparrow$) & 3.1E-6 ($\uparrow$) \\ \hline
\end{tabular}
\end{table}

\begin{figure*}
    \centering
    \includegraphics[width=0.83\linewidth]{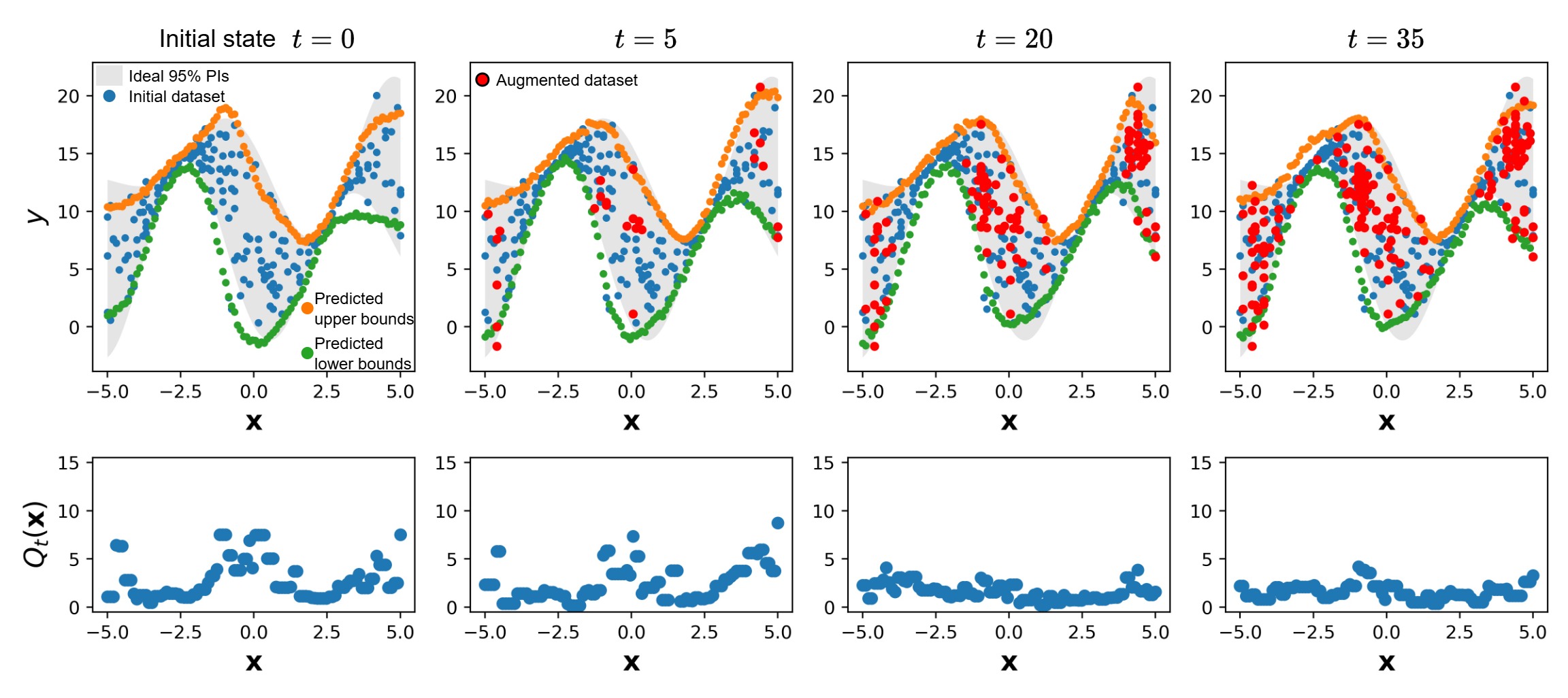}
    \caption{Adaptive sampling process using ASPINN on the \texttt{cos} problem.}
    \label{fig:ex_cos}
\end{figure*}

\begin{figure*}
    \centering
    \includegraphics[width=0.83\linewidth]{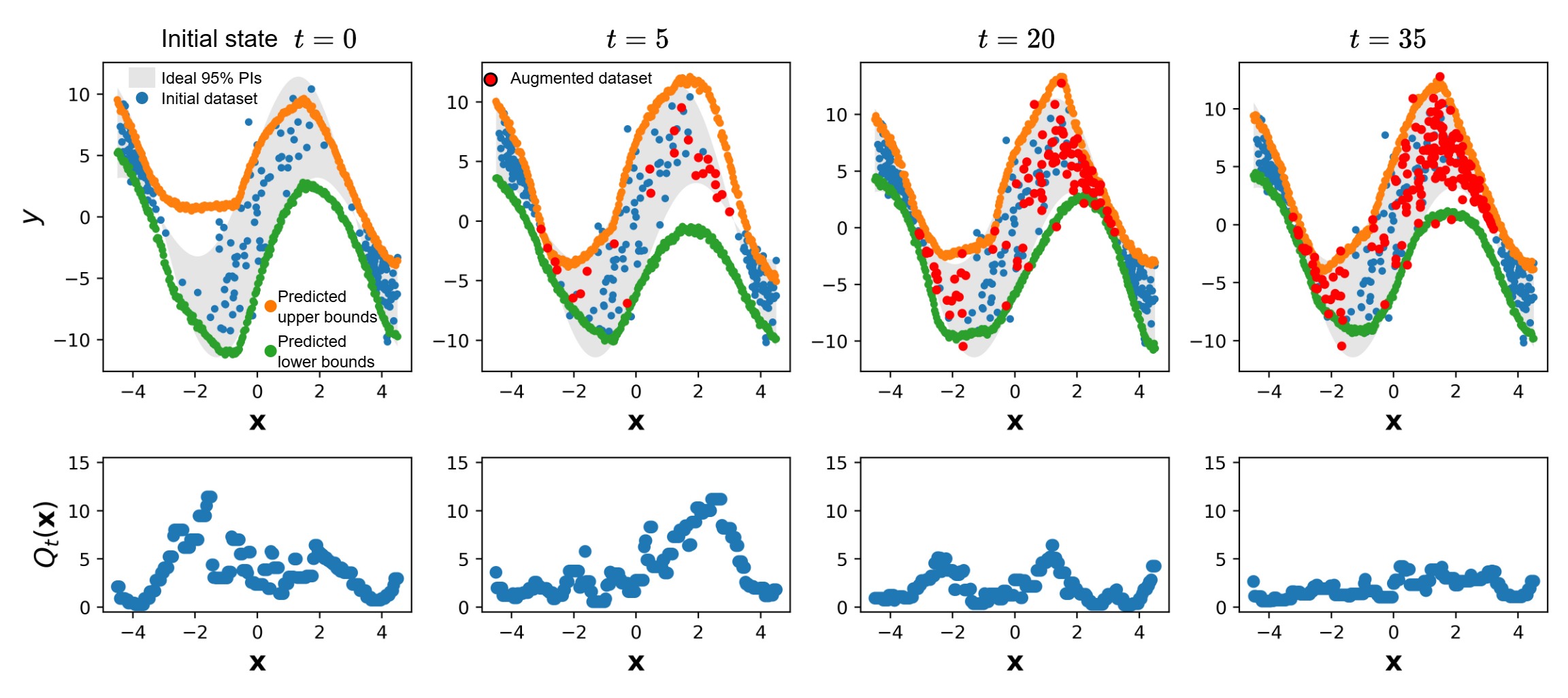}
    \caption{Adaptive sampling process using ASPINN on the \texttt{hetero} problem.}
    \label{fig:ex_hetero}
\end{figure*}

\begin{figure*}
    \centering
    \includegraphics[width=0.83\linewidth]{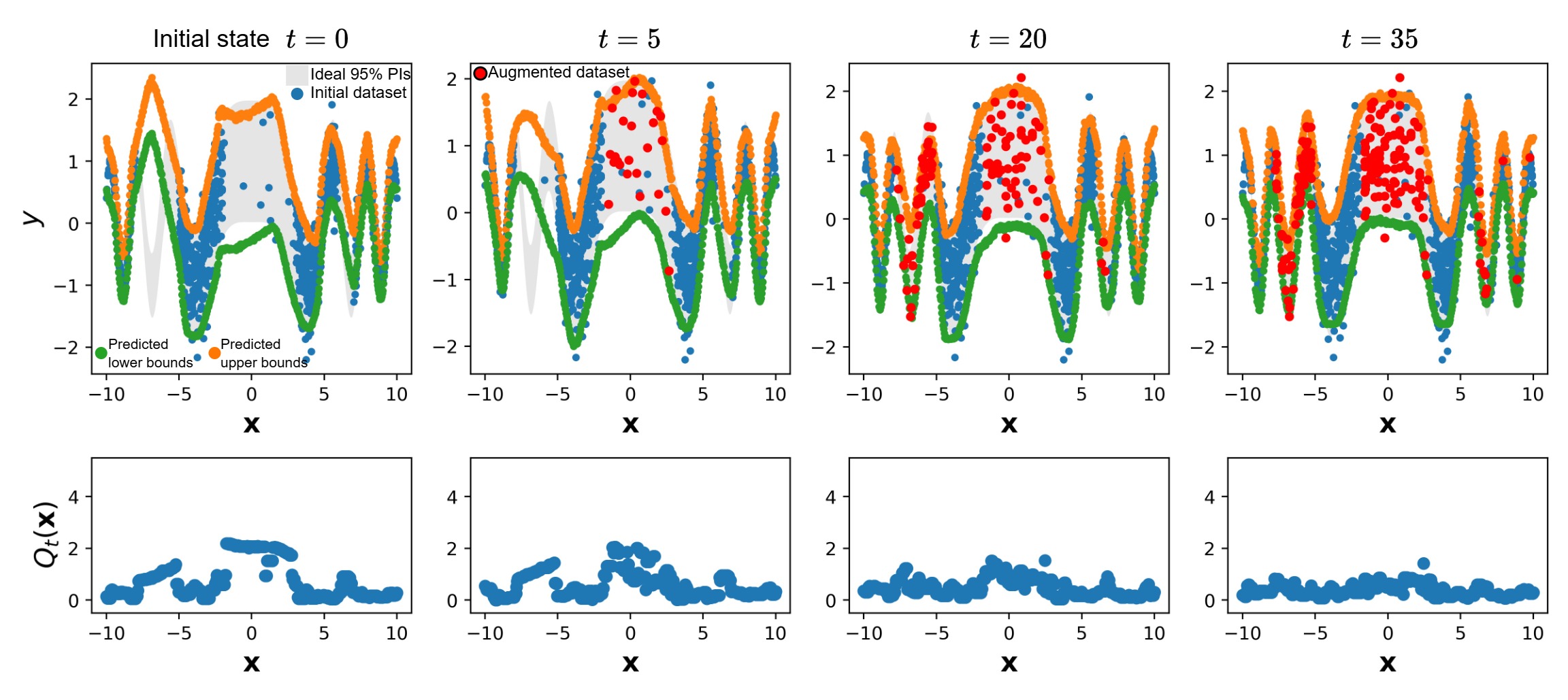}
    \caption{Adaptive sampling process using ASPINN on the \texttt{cosqr} problem.}
    \label{fig:ex_cosqr}
\end{figure*}

Furthermore, we illustrate some of the results obtained by ASPINN for problems \texttt{cos}, \texttt{hetero}, and \texttt{cosqr} in Figures~\ref{fig:ex_cos}, \ref{fig:ex_hetero}, and \ref{fig:ex_cosqr}, respectively.
The figures show the problems' initial state and the augmented datasets obtained during iterations $t=5$, $t=20$, and $t=35$.
They also display the corresponding calculated potential epistemic uncertainty $Q_t(\mathbf{x})$ for all values of the input domain.

\subsection{Experiments with Simulated Field Data} \label{sec:MZapp}

It is important to note that AS techniques are applied in contexts where experts can perform experiments in some selected locations of the input domain. 
Thus, candidate sampling locations are limited. For example, besides the 1-D problems presented by \citet{NFs}, they considered a multi-dimensional problem called Pendulum. 
This problem consists of four input variables with one of them, torque, representing the action variable (i.e., the agent's only control is over the torque applied). 
In the same fashion as the Pendulum problem, we presented experiments for a multi-dimensional problem (i.e., four inputs) simulating a real-world agricultural site, where the action variable is given by the fertilizer rate applied.
However, our problem is more complex than Pendulum.

The behavior of this site is modeled based on a symbolic skeleton that was extracted from an actual winter wheat field
~\cite{MultiSetSR}:
\begin{equation}
    \label{eq:skeleton}
   y=c_1 + c_2\, \texttt{tanh}(c_3 + c_4\, \mathbf{x}^{Nr}), 
\end{equation}
where $c_1$--$c_4$ are placeholder constants that may depend on other explanatory variables.

To facilitate reference, we reproduce the yield function considered in this work:
\begin{equation}
    \label{eq:function}
    y=f(\mathbf{x}) = \frac{\mathbf{x}^{P}}{15} + \left(\frac{\mathbf{x}^{A}}{\pi} + 1\right) \tanh\left(\frac{0.1 \, \mathbf{x}^{Nr}}{3 \mathbf{x}^{VH} + 2}\right) + \varepsilon_a(\mathbf{x}),
\end{equation}
where $\mathbf{x} = [\mathbf{x}^{P}, \mathbf{x}^{A}, \mathbf{x}^{VH}, \mathbf{x}^{Nr}]$ comprises the following site-specific covariates: annual precipitation (mm), terrain aspect (radians), Sentinel-1 backscattering coefficient from the Vertical Transmit-Horizontal Receive Polarization band, and applied N rate (lbs/ac), respectively.
The aleatoric noise is modeled as $\varepsilon_a(\mathbf{x}) = \mathcal{N}(0, (\mathbf{x}^{P} + \mathbf{x}^{Nr}) / 1500)$.

We considered $\mathbf{x}^{P} \in [75, 150]$, $\mathbf{x}^{A} \in [\pi/4, \pi/2]$, $\mathbf{x}^{VH} \in [0.5, 1]$, and $\mathbf{x}^{Nr} \in [0, 30, 60, 90, 120, 150]$.
These values were selected to reflect realistic conditions based on past observations from the sub-region of the field used to model our simulated field site.
During each iteration of the process, we sample a new precipitation value such that $\mathbf{x}_t^{P} \sim \mathcal{U}(75, 150)$.
Similarly, we accounted for variations in the terrain aspect by modeling $\mathbf{x}_t^{A}$ as $\mathcal{U}(\pi/4, \pi/2)$.
This assumption reflects slight alterations in the landscape each growing season, influenced by factors such as weather conditions and the use of heavy machinery.

Variable $\mathbf{x}^{VH}$ is associated with soil moisture content, where lower values correspond to drier soil conditions.
Given the absence of additional variables to model soil moisture accurately and since this is beyond the scope of our study, we developed a simplified moisture function that incorporates precipitation and terrain aspect.
In particular, we consider $\mathbf{x}_t^{VH} = \frac{\mathbf{x}_t^{P}}{150} \mathbf{x}_t^{A}$, reflecting that higher precipitation and terrain aspect values result in greater soil moisture content. 
Based on this parameterization, the initial dataset $\mathbf{X}_{obs}^{(t=0)}$ is generated by randomly sampling 50 data points, each representing a distinct growing season.

Here, we justify the selection of these underlying and noise functions.
From comparing Equations~\ref{eq:skeleton} and \ref{eq:function}, it is observed that the constant placeholders were assigned the following values: $c_1=\frac{\mathbf{x}^{P}}{15}$, $c_2=(\frac{\mathbf{x}^{A}}{\pi} + 1)$, $c_3=0$, and $c_4=(\frac{0.1}{3 \mathbf{x}^{VH} + 2})$.
Below, we analyze each of these expressions.
It is important to clarify that our goal is not to derive precise functional expressions for the coefficients $c_1$--$c_4$ in order to model the underlying function of the field accurately.
Rather, our aim is to design a yield function that exhibits behavior consistent with agronomic principles, informed by past observations of an actual field.

In a previous work
~\cite{rcurves2023},
we utilized counterfactual explanations to analyze the influence of a set of ``passive features" over the shape of the response curves generated for the response variable and a selected ``active feature."
In the context of this work, the agricultural field site represents a multivariate system.
We are interested in the analysis of nitrogen-yield response (N-response) curves.
N-response curves are tools that allow for the analysis of the site-specific responsivity to all admissible values of the N fertilizer rate, which serves as the selected ``active feature."
Nevertheless, the shape of N-response curves may be influenced not only by the relationship between the response variable and the active feature but also by other factors, termed "passive features."
In this case, we consider the variables $\mathbf{x}^{P}$, $\mathbf{x}^{A}$, and $\mathbf{x}^{VH}$ as passive features.

In
~\cite{rcurves2023},
we studied an early-yield prediction dataset of winter wheat.
The findings indicate that, although precipitation $\mathbf{x}^{P}$ is a critical factor for crop production, it has minimal impact on N responsivity.
This suggests that $\mathbf{x}^{P}$ is independent of the other features and only shifts the N-response curves vertically without altering their shape.
In Eq.~\ref{eq:skeleton}, $c_1$ acts as an independent term responsible for vertical shifts, which is why it is modeled as a function of $\mathbf{x}^{P}$.
In addition, variables $\mathbf{x}^{A}$ and $\mathbf{x}^{VH}$ were identified as having a significant impact on the shape of N-response curves, making them key factors in this study.

Furthermore, $c_2$ stretches the N-response curves vertically.
We argue this behavior corresponds to that of the terrain aspect $\mathbf{x}^{A}$ (i.e., the slope orientation).
In terrain with varying elevations located in the Northern Hemisphere, regions that are facing north and east have limited sunlight during the day and are more prone to snow retention.
These are factors that may affect the responsiveness of the fertilizer.
For instance, we observed that regions facing north ($\mathbf{x}^{A} = 0$) correspond to flatter N-response curves than those facing south ($\mathbf{x}^{A}=\pi$).
Our simulated field site is being modeled as a field site that is located within a sub-region of an actual field whose $\mathbf{x}^{A}$ values vary between $\pi/4$ and $\pi/2$.
Within this sub-region, we found that considering $c_2=(\frac{\mathbf{x}^{A}}{\pi} + 1)$ adjusts reasonably well to the variation in vertical stretching of the estimated N-reponse curves.

Coefficient $c_3$ causes horizontal shifts, which are not observed in the estimated N-response curves obtained for the studied area.
Hence, for the sake of simplicity, we select $c_3$ to be equal to 0.
Finally, $c_4$ controls the horizontal stretching of the curve.
A lower $c_4$ value causes the output of the function to increase more gradually as $\mathbf{x}$ increases.
Conversely, a higher value of $c_4$ leads to a steeper increase, causing the function to reach its saturation point more rapidly.
Dry soil has a lower capacity to retain and absorb nutrients and, thus, reaches the saturation point more quickly than moist soil when applying N fertilizer.
Therefore, we model $c_4$ as an inverse function of $\mathbf{x}^{VH}$: $c_4=(\frac{0.1}{3 \mathbf{x}^{VH} + 2})$.

On the other hand, the heteroskedastic aleatoric noise is modeled as $\varepsilon_a(\mathbf{x}) = \mathcal{N}(0, (\mathbf{x}^{P} + \mathbf{x}^{Nr}) / 1500)$.
This is based on the observations that both precipitation $\mathbf{x}^{P}$ and N fertilizer rate $\mathbf{x}^{Nr}$  contribute to variability and uncertainty in agricultural yield outcomes.
That is, increased precipitation can lead to greater variability in soil conditions, such as runoff, which in turn affects nutrient availability and crop health.
In addition, the effects of the N fertilizer do not always increase monotonically. 
High levels of nitrogen can lead to diminishing returns or even negative effects, such as nutrient imbalances or environmental stress on the plants. 
These unpredictable responses add to the uncertainty in yield, particularly at higher N fertilizer rates.

Finally, we assess the differences in AUUC values achieved by ASPINN across the ten AS iterations compared to those obtained by the other methods.
Table~\ref{tab:MZstats} reports the $p$-values from the paired $t$-tests comparing ASPINN with the alternative approaches.
The results indicate that the differences in AUUC values are statistically significant (i.e., $p$-value $<0.05$).

\begin{table}[t]
\small
\centering
\caption{Statistical significance tests --- Simulated field site. $p$-values obtained comparing ASPINN to the other methods.}
\label{tab:MZstats}
\def\arraystretch{1.5}
\begin{tabular}{|c|c|}
\hline
\textbf{\begin{tabular}[c]{@{}c@{}}Compared\\ Method\end{tabular}} & $p$-value \\ \hline
NF-Ensemble & 1.3E-4 ($\uparrow$) \\ \hline
GP & 4.4E-2 ($\uparrow$) \\ \hline
MC-Dropout & 6.7E-3 ($\uparrow$) \\ \hline
\end{tabular}
\end{table}

\end{document}